\documentclass[acmsmall,screen,nonacm]{acmart}
\usepackage{tikz}
\usetikzlibrary{shapes, arrows, positioning, fit, calc, shadows}
\usepackage{graphicx}
\usepackage{booktabs}
\usepackage{array}
\usepackage{tabularx}
\usepackage{amsmath}
\usepackage[table]{xcolor}
\newcommand{\ATD}{\textit{Agentic Technical Debt}}
\newcommand{\ST}{\textit{Stochastic Tax}}

\AtBeginDocument{}
\setcopyright{none}
\copyrightyear{2026}
\acmYear{2026}
\acmDOI{}

\begin{document}

\title[Governing Technical Debt in Agentic AI Systems]{Governing Technical Debt in Agentic AI Systems}

\author{Muhammad Zia Hydari}
\authornote{All authors contributed equally to this research.}
\affiliation{%
  \institution{School of Business, University of Pittsburgh}
  \city{Pittsburgh}
  \state{Pennsylvania}
  \country{USA}}
\email{hydari@alum.mit.edu}

\author{Raja Iqbal}
\authornotemark[1]
\affiliation{%
  \institution{Ejento AI}
  \city{Seattle}
  \state{Washington}
  \country{USA}}
\email{raja@ejento.ai}

\author{Narayan Ramasubbu}
\authornotemark[1]
\affiliation{%
  \institution{School of Business, University of Pittsburgh}
  \city{Pittsburgh}
  \state{Pennsylvania}
  \country{USA}}
\email{narayanr@pitt.edu}

\renewcommand{\shortauthors}{Hydari, Iqbal, and Ramasubbu}

\begin{abstract}
Agentic AI systems are increasingly being explored as production infrastructure: they reason over multiple steps, call tools, act through workflows, and adapt through memory and feedback. These systems create governance challenges that are not fully captured by traditional software or predictive ML technical debt. We define \ATD{} as the accumulated liability created when prompts, memory, tool schemas, orchestration graphs, control policies, and observability routines are patched together faster than they can be validated, standardized, and governed. We define \ST{} as the recurring operating burden of keeping probabilistic agent behavior within acceptable bounds. The distinction matters: debt is a stock of design and governance liability, while the tax is a flow of operating cost that arises because stochastic agents act through tools and workflows. We outline how managers can make both visible through lightweight dashboards and governance controls.
\end{abstract}

\maketitle

\section{Introduction}

Technical debt describes the long-term costs of short-term design or implementation decisions \cite{cunningham1992wycash}. In conventional software, debt is often structural: modularity erodes, tests lag behind features, and maintenance costs rise, but failures remain comparatively reproducible \cite{ramasubbu2015managing}. In ML, debt is more systemic: pipelines entangle data, models, and external signals in ways that create hidden coupling and silent degradation \cite{sculley2015hidden}. Agentic AI introduces a third setting. Going beyond predicting or classifying, these systems can plan, call tools, observe outcomes, and revise behavior across multi-step workflows.

The managerial problem is that stochastic output is now coupled to delegated action. Prompts can determine which tool is called, which record is changed, which exception is escalated, or which customer receives a decision. In this context, we define \textbf{\ATD{}} as the accumulated liability that emerges when prompts, memory representations, tool schemas, orchestration graphs, control policies, and governance workflows are patched together faster than they can be validated, standardized, and made reliably auditable. We define \textbf{\ST{}} as the recurring operating burden of keeping stochastic agent behavior within acceptable bounds through evaluation, monitoring, gating, retries, escalation, revalidation, latency management, token and context processing, and guardrail maintenance.

\ATD{} and \ST{} are related but distinct. Debt is an accumulated stock of design and governance liability. Tax is a recurring flow of operating burden. Debt can amplify the tax, but some tax remains even when debt is absent because agentic systems are stochastic and act through tools, memory, and workflows. The term \ST{} therefore has two meanings: it is a tax incurred because the agentic system is stochastic, and the tax itself is stochastic because its rate varies with user adoption, surface area, workflow length, autonomy, model behavior, and accumulated debt.

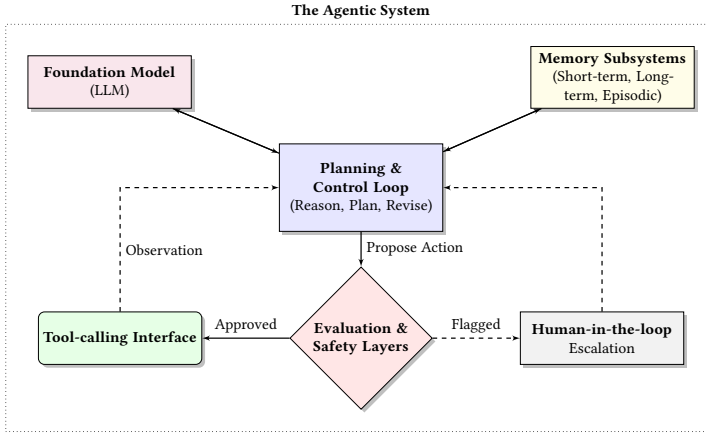
\begin{figure}[h]
\centering
\resizebox{0.68\columnwidth}{!}{\begin{tikzpicture}[
    node distance=0.8cm and 2cm,
    block/.style={rectangle, draw, fill=white, text width=3.5cm, minimum height=1.2cm, align=center, drop shadow},
    decision/.style={diamond, draw, fill=red!10, text width=2.5cm, align=center, inner sep=0pt, drop shadow},
    line/.style={draw, -latex', thick},
    dashedline/.style={draw, dashed, -latex', thick}
]

% Central Node: Planning & Control
\node [block, fill=blue!10, minimum height=2cm] (control) {\textbf{Planning \& Control Loop}\\(Reason, Plan, Revise)};

% Top Nodes: Intelligence and State
\node [block, above left=of control, fill=purple!10] (llm) {\textbf{Foundation Model}\\(LLM)};
\node [block, above right=of control, fill=yellow!10] (memory) {\textbf{Memory Subsystems}\\(Short-term, Long-term, Episodic)};

% Bottom Node: Safety
\node [decision, below=of control] (safety) {\textbf{Evaluation \& Safety Layers}};

% Execution Nodes
\node [block, left=of safety, fill=green!10, rounded corners] (tools) {\textbf{Tool-calling Interface}};
\node [block, right=of safety, fill=gray!10] (human) {\textbf{Human-in-the-loop}\\Escalation};

% Connections
\path [line] (control) -- (llm);
\path [line] (llm) -- (control);
\path [line] (control) -- (memory);
\path [line] (memory) -- (control);
\path [line] (control) -- node[right] {Propose Action} (safety);
\path [line] (safety) -- node[above] {Approved} (tools);
\path [dashedline] (safety) -- node[above] {Flagged} (human);
\draw [dashedline] (tools.north) |- node[near start, right] {Observation} (control.west);
\draw [dashedline] (human.north) |- (control.east);

% Grouping Box
\node[draw, dotted, inner sep=0.5cm, fit=(control) (llm) (memory) (safety), label=above:\textbf{The Agentic System}] (system) {};

\end{tikzpicture}}
\caption{Reference architecture of an agentic AI system.}
\Description{Reference architecture showing a foundation model connected to planning, memory, tools, evaluation, guardrails, and escalation.}
\label{fig:agentic_arch}
\end{figure}

In production, an agentic system typically combines a foundation model, a planning and control loop, a tool-calling interface, memory, evaluation and safety layers, and human escalation paths \cite{yao2022react,bommasani2021opportunities}. Because the loop conditions on stochastic model outputs and persistent state, small changes can alter action sequences, complicating reproducibility, auditability, and change control. Figure~\ref{fig:agentic_arch} summarizes a common reference architecture.

This architecture accelerates debt accumulation because teams often prototype by patching prompts, adding wrappers, and serializing handoffs before interfaces and policies are stabilized. Each local fix can increase coupling among prompts, tool schemas, memory representations, and safety checks, raising the cost of future changes. As organizations compose agents, tools, and data products into larger workflows, each additional interface expands the surface for brittle orchestration, weak observability, and recurring evaluation costs.

\section{Understanding Debt and Tax}

To manage risk, we must distinguish \ATD{} from earlier debt paradigms and from \ST{}. Table~\ref{tab:debt} compares the three debt paradigms across what accumulates, how coupling appears, and what managers can do.

\begin{table}[h]
\caption{Comparison of Technical Debt Paradigms}
\label{tab:debt}
\footnotesize
\begin{tabularx}{\textwidth}{p{0.16\textwidth}p{0.25\textwidth}X p{0.23\textwidth}}
\toprule
\textbf{Debt paradigm} & \textbf{What accumulates} & \textbf{Coupling and symptom} & \textbf{Managerial response} \\
\midrule
Structural software debt & Code complexity, poor modularity, weak tests & Module dependencies create deterministic defects and slow feature velocity. & Refactor code, improve modularity, expand regression tests. \\
Predictive ML debt & Data dependencies, feature entanglement, pipeline fragility & Coupling among data, features, models, and signals creates silent degradation and drift. & Version data and models, prune features, monitor drift, retrain carefully. \\
Agentic AI debt & Prompts, memory, tool schemas, orchestration graphs, policies, monitoring gaps & Coupling among stochastic outputs, tools, memory, actions, and rules creates action variance, malformed tool calls, retry loops, tail latency, and escalation spikes. & Standardize tool contracts, test traces, govern memory, redesign workflows, tune autonomy. \\
\bottomrule
\end{tabularx}
\end{table}

In traditional software, the same input follows the same rule and produces the same behavior; if the code is messy, it is predictably messy \cite{fowler1999refactoring}. Predictive ML debt is less visible because the logic is learned from data and can degrade silently \cite{sculley2015hidden}. Agentic debt extends this concern to action. The system chooses tools, sequences steps, writes or retrieves memory, invokes policies, and may trigger changes in external systems. The user's intent is stated at a high level, while the agent's execution path is generated probabilistically across technical and organizational constraints.

\ATD{} accumulates through five recurring mechanisms that managers can diagnose:
\begin{itemize}
    \item \textbf{Autonomy:} agents exercise delegated authority by selecting tools and sequencing steps, creating a principal-agent gap between intent and execution.
    \item \textbf{Semantic ambiguity:} instructions are written in natural language rather than formal syntax, making small wording changes behaviorally consequential.
    \item \textbf{Stochasticity:} identical inputs can yield different plans, tool calls, or intermediate traces.
    \item \textbf{Persistent state:} memory improves continuity but creates liabilities when stale, inconsistent, or poorly governed.
    \item \textbf{Latency amplification:} suboptimal LLM calls can cost seconds, cascade into retries, and degrade tail performance.
\end{itemize}

A prompt with conflicting instructions, an unversioned connector, or a brittle routing rule is debt when it makes future change, validation, or control harder than it should be. \ST{}, by contrast, is the recurring burden of managing stochastic operation. It includes healthy control cost as well as avoidable burden amplified by debt. A well-governed agentic workflow may still require evaluation and monitoring; a debt-laden one will require more retries, escalations, trace reviews, and emergency controls. Table~\ref{tab:atd_st_comparison} summarizes this boundary.

\begin{table}[h]
\caption{Comparing \ATD{} and \ST{}}
\label{tab:atd_st_comparison}
\scriptsize
\begin{tabularx}{\textwidth}{p{0.18\textwidth}X X}
\toprule
\textbf{Dimension} & \textbf{\ATD{}} & \textbf{\ST{}} \\
\midrule
Basic idea & Accumulated liability from expedient design and governance choices & Recurring cost of operating stochastic agent behavior safely \\
Time pattern & Builds up as a stock & Paid repeatedly as a flow \\
Where it appears & Prompts, memory, tool schemas, orchestration, policies, monitoring gaps & Evaluation, monitoring, retries, escalation, revalidation, latency, token/context cost, guardrails \\
Main question & What shortcuts make future change harder? & What recurring burden keeps behavior acceptable? \\
Managerial response & Refactor, standardize, document, redesign, create contracts & Measure, budget, automate checks, set thresholds, triage high-tax workflows \\
\bottomrule
\end{tabularx}
\end{table}

A lightweight dashboard can make this visible. Let $\overline{\mathrm{ST}}_{w,t}$ denote the average stochastic tax per completed transaction for workflow $w$ during period $t$:
\[
\overline{\mathrm{ST}}_{w,t}=\frac{C^{\mathrm{eval}}_{w,t}+C^{\mathrm{monitor}}_{w,t}+C^{\mathrm{retry}}_{w,t}+C^{\mathrm{escalate}}_{w,t}+C^{\mathrm{revalidate}}_{w,t}+C^{\mathrm{latency}}_{w,t}+C^{\mathrm{token}}_{w,t}+C^{\mathrm{security}}_{w,t}}{N_{w,t}}.
\]
Here $N_{w,t}$ is the number of completed transactions. The numerator captures recurring evaluation, monitoring, retry and repair, human escalation, revalidation after model, tool, prompt, policy, or context changes, latency-related delay, token and context-processing cost, and security or guardrail maintenance. This is not a universal metric; it is a managerial dashboard for asking whether the burden of a workflow is stable, rising, or concentrated in particular channels. For a fuller operationalization of this dashboard view, see \citet{hydari2026modeling}.

\section{Debt Accumulation}

\begin{figure}[h]
\centering
\resizebox{0.78\columnwidth}{!}{\begin{tikzpicture}[
    scale=0.5, 
    transform shape,
    node distance=1.0cm and 1.0cm,
    agent/.style={rectangle, draw=black!60, fill=blue!10, rounded corners, drop shadow, minimum height=1cm, minimum width=2.5cm, align=center, font=\sffamily\small},
    cloud/.style={circle, draw=red!80, fill=red!10, dashed, drop shadow, minimum size=2.5cm, align=center, text=red!80, font=\sffamily\bfseries\small},
    guardrail/.style={rectangle, draw=black!60, dashed, fill=yellow!10, drop shadow, minimum height=1.5cm, minimum width=5cm, align=center, font=\sffamily\small},
    line/.style={draw, -latex', thick, color=black!70},
    retry/.style={draw, dotted, -latex', color=red!80, thick}
]

% Nodes
\node [agent] (intake) {Intake Agent};
\node [agent, right=of intake] (vendor) {Vendor Check};
\node [agent, right=of vendor] (policy) {Policy Check};
\node [agent, right=of policy] (execute) {Execute};
\node [cloud, above=of vendor, xshift=-0.5cm] (retry) {Retry Loop};
\node [guardrail] (guards) at (execute |- retry) {\textbf{Guardrails}\\Validate, Gate, Monitor};

% Connections
\path [line] (intake) -- (vendor);
\path [line] (vendor) -- (policy);
\path [line] (policy) -- (execute);
\path [retry] (intake.north) edge [bend left=45] (retry.west);
\path [retry] (vendor.north) edge (retry.south);
\path [retry] (policy.north) edge [bend right=45] (retry.east);
\path [retry] (retry.north) edge [bend right=90] node[midway, above, font=\tiny, color=red] {Correction} (intake.north);
\draw [dashed, ->, thick, color=black!60] (policy.north) edge [bend left=20] (guards.west);
\draw [dashed, ->, thick, color=black!60] (execute.north) -- (guards.south);

\end{tikzpicture}}
\caption{The sequential trap: serial agent handoffs create latency, retries, and recurring \ST{}.}
\Description{Sequential agent workflow with intake, vendor check, policy check, execution, retry loop, and guardrails.}
\label{fig:sequential_trap}
\end{figure}
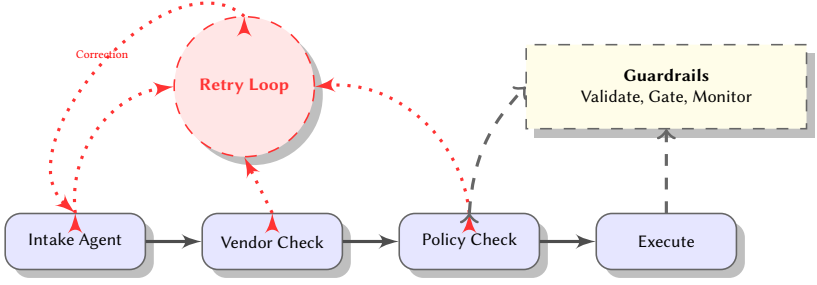

\textbf{The sequential trap.} One severe form of \ATD{} is the \textit{Orchestration Jungle}. In a real-world insurance workflow observed by one of the authors, specialized agents for auto, home, and life policies were orchestrated sequentially (Figure~\ref{fig:sequential_trap}). The design was logically sound, but it created latency and change debt: each agent added delay, each handoff introduced another failure point, and changes to one step could require revalidating the chain. This is the agentic analogue of the \textit{Pipeline Jungle}~\cite{sculley2015hidden}: a workflow that functions but resists modification. In such cases, the economical fix may not be another prompt patch or wrapper, but redesigning the workflow as a parallelized directed acyclic graph.

\textbf{Prompt cascades.} \citet{sculley2015hidden} warned of correction cascades in ML. Agentic systems produce prompt cascades: teams add local instructions such as ``be polite,'' ``do not promise refunds,'' or ``respond in JSON'' until the system prompt becomes hard to edit safely. The corrective move is not always more prompting. Some constraints should move into schemas, policy services, deterministic checks, or smaller prompts with explicit tests.

\textbf{Dependency drift.} In generative AI, the organization rents part of the intelligence. Homogenization around a few foundation models creates single points of failure and can undermine resilience \cite{bommasani2021opportunities}. A prompt or tool call that performed reliably before a model, safety-policy, or vendor update may behave differently after it. The enterprise has accrued agentic debt not because its code rotted, but because the foundation beneath the application shifted.

\section{Operationalizing Governance}

We cannot completely remove stochasticity or autonomy from agentic systems, but we can control their cost. Governance should define acceptable behavior, detect regressions before release, and constrain action authority at runtime. Table~\ref{tab:governance_mapping} maps controls to debt mechanisms and tax signals.

\begin{table}[h]
\caption{Mapping Governance Controls to Debt Mechanisms and \ST{} Signals}
\label{tab:governance_mapping}
\footnotesize
\begin{tabularx}{\textwidth}{p{0.23\textwidth}p{0.27\textwidth}p{0.24\textwidth}X}
\toprule
\textbf{Governance control} & \textbf{Debt mechanism addressed} & \textbf{\ST{} signal} & \textbf{Question} \\
\midrule
Golden-set evaluation and trace-level diffing & Semantic ambiguity, run-to-run variance & Policy violations, action variance, regression failures & Are behavior changes detected before release? \\
Tool schema contracts and deterministic checks & Fragile tool interfaces and malformed actions & Retry rate, schema failure rate & Are tool failures prevented before the model retries? \\
Model gateway and versioned registry & Dependency drift and provider coupling & Revalidation effort after model or vendor changes & Can the organization absorb model changes without rewrites? \\
Graduated autonomy and policy permissions & Principal-agent gap from delegated authority & Escalation rate, override rate, blast-radius incidents & Which actions may the agent propose, and which may it execute? \\
Workflow graph redesign and parallelization & Orchestration jungle and latency amplification & P95 latency, retry loops, exception handling & Is the workflow structure creating avoidable burden? \\
\bottomrule
\end{tabularx}
\end{table}

First, move from ``vibes" to evaluated traces. Agent workflows require deterministic checks, including schemas, tool contracts, and policy rules; a versioned golden dataset of critical intents and edge cases; and judge models or human reviewers that score traces on task success, policy compliance, and evidence faithfulness. Deploy changes only when scores remain within thresholds and failures produce trace-level diffs.

Second, abstract the intelligence layer. A model gateway can standardize prompts, tool schemas, and response formats and route across models without application rewrites. Pair it with an asset registry for agents, prompts, and tools; a version-controlled policy service enforced at runtime; and end-to-end observability for tool calls and outcomes. This does not eliminate revalidation, but it prevents every model update from becoming an application rewrite.

Third, make autonomy configurable. Use proposer mode for high-risk actions, conditional autonomy within registered limits, and full autonomy only for low-blast-radius tasks. A simple traffic-light policy can make the boundary operational: \textbf{green} actions, such as classifying or routing support tickets, may run autonomously because they are reversible and low impact; \textbf{yellow} actions, such as scheduling accounts-payable transactions under registered dollar and vendor-identity thresholds, may proceed only within explicit limits and otherwise downgrade to proposer mode; and \textbf{red} actions, such as writing to production databases or changing ERP master data, remain human-authorized by default. When signals cross thresholds, the system can automatically downgrade autonomy, route to a more reliable model, or trigger a circuit breaker until revalidation completes.

\section{Conclusion}

Agentic AI shifts engineering from specifying deterministic logic to governing probabilistic action. Organizations that treat agents as ordinary software components may accumulate \ATD{} and pay a rising \ST{} through retries, monitoring overhead, and emergency guardrails. The aim is not to eliminate stochasticity, but to make its recurring burden visible, bounded, and governed. With disciplined evaluations, model gateways, versioned registries, and graduated autonomy, enterprises can scale agentic systems while keeping behavior auditable, failures containable, and operating costs predictable.

\end{document}